\documentclass[conference]{IEEEtran}

\usepackage{cite}
\usepackage{url}
\usepackage{amsmath,amssymb,amsfonts}
\usepackage{algorithmic}
\usepackage{graphicx}
\usepackage{subcaption}
\usepackage{textcomp}
\usepackage{xcolor}
\usepackage{hyperref}

\def\BibTeX{{\rm B\kern-.05em{\sc i\kern-.025em b}\kern-.08em
    T\kern-.1667em\lower.7ex\hbox{E}\kern-.125emX}}

\usepackage[all]{nowidow}
\usepackage{multirow}
\usepackage{array}
\usepackage{booktabs}
\newcolumntype{P}[1]{>{\centering\arraybackslash}p{#1}}

\begin{document}

\title{Green Prompting: Characterizing Prompt-driven Energy Costs of LLM Inference}

\author{
    \IEEEauthorblockN{
        Marta Adamska\IEEEauthorrefmark{1}, 
        Daria Smirnova\IEEEauthorrefmark{1}, 
        Hamid Nasiri\IEEEauthorrefmark{1}, 
        Zhengxin Yu\IEEEauthorrefmark{1}, 
        and Peter Garraghan\IEEEauthorrefmark{1}
    }
    \IEEEauthorblockA{\IEEEauthorrefmark{1}School of Computing and communications, Lancaster University, UK \\
    Email: \{m.adamska, d.smirnova1, h.nasiri, z.yu8, p.garraghan\}@lancaster.ac.uk}
}

\maketitle

\begin{abstract}
Large Language Models (LLMs) have become widely used across various domains spanning search engines, code generation, and text creation. However, the high cost of inference is a major concern associated with their adoption, as it impacts both their financial feasibility and, more importantly, sustainability. In this study, we empirically examine how different prompt and response characteristics impact LLM inference energy cost. We conduct experiments leveraging three open-source transformer-based LLMs across three task types--question answering, sentiment analysis, and text generation. For each inference, we analyzed prompt and response characteristics (length, semantic meaning, latency, energy consumption). \\
\indent Our results demonstrate that even when presented with identical tasks, models generate responses with varying characteristics and subsequently exhibit distinct energy consumption patterns. We found that prompt length is less significant than the semantic meaning of the task itself. In addition, we identified specific keywords associated with higher or lower energy usage that vary between associated tasks, with certain keywords seeing up to 60\% energy reduction. These findings highlight the importance of prompt design in optimizing inference efficiency. We conclude that the semantic meaning of prompts and certain task-related keywords significantly impact inference costs, leading the way for deeper exploration towards sustainable LLM operation.

\end{abstract}


\section{Introduction}
\label{introduction}
Transformer models have revolutionized the AI landscape by enabling faster and more accurate predictions through their innovative use of self-attention mechanisms \cite{vaswani2017attention}. 
The transformer architecture serves as the underlying architecture for most modern Large Language Models (LLMs) and, unlike traditional neural networks, they overcome key limitations by making use of multiple attention mechanisms and processing large datasets, which allows to efficiently capture complex relationships within data. This architecture allows the use of LLMs for linguistic tasks ranging from medical applications \cite{thirunavukarasu2023large}, search engines \cite{microsoft2025bing}, to enhanced code development \cite{zeng2022extensive,bubeck2023sparks}. 

Although the versatility and adoption of LLMs can provide benefits, their widespread use raises significant concerns, particularly regarding energy consumption and environmental impact. The carbon footprint and energy usage of an LLM life cycle vary greatly depending on factors such as the geographical location of the data centers and the hardware on which they run\cite{strubell2020energy}. Despite that, the growing demand for more complex and large-scale deployments suggests that the demand for energy for AI systems will grow \cite{iea2024electricity}.
There has been a lot of attention paid to the AI training cost. For example, BLOOM -- a 176B parameter model -- consumed 433,196 kWh \cite{luccioni2023estimating}. Another large energy consumer is inference, which uses the trained model to generate outputs from previously unseen data inputs. For comparison, it is estimated that the energy needed for a single inference for BLOOM was on average 3.96 Wh. Although the energy cost of individual inference is small, the inference stage can make up to 65\% \cite{wu2022sustainable} of the carbon footprint of an AI model's life cycle due to billions of inference calls, responsible for significant energy costs.

Previous studies examine different factors contributing to LLM energy consumption during inference across model architecture, hardware accelerators \cite{khan2021npe,armeniakos2022hardware}, parallelism and batching \cite{patel2024splitwise, agrawal2024taming}, and workload types \cite{hu2024characterization, stojkovic2024towards}. Less attention has been given to how prompts and their characteristics impact the energy usage of LLM inference. Prompt engineering, the practice of designing prompts to achieve the desired output, has primarily been used to achieve the best possible accuracy \cite{arora2022ask, cheng2023black, vatsal2024survey}. However, the role of prompt optimization in reducing energy consumption of the LLMs remains an open question \cite{luccioni2024power, luccioni2023estimating}.

While various approaches have been explored to address the challenge of energy consumption during LLM inference, the potential of prompt engineering, different qualities of prompts and responses, in this context remain largely unexplored. To identify the patterns behind energy usage, it is crucial to analyze the impact of different prompt characteristics and determine the key drivers of energy consumption. This issue is particularly significant as  prompt characteristics can vary widely within LLM-based systems. 

In this paper, we present an empirical study of LLMs to understand their inference energy usage. The studied models are based on transformer architecture and are similar in parameter size; some of the differentiating qualities include vocabulary size, number of hidden layers and training data. In order to understand the energy patterns in those models, we ran multiple prompts with different qualities, multiple tasks, and changing hyperparameters. Each model received the same set of prompts with the same hyperparameter settings. Energy usage and time taken to respond (latency) were measured during each of the inference runs. 

The results provided insight into which features impact energy consumption. Our findings demonstrated that the length of the response generated by the LLM is a major driver of energy consumption. Experiments also suggest that response correctness does not significantly influence the energy consumption. All results are discussed in Section \ref{sec:analysis}. This insight opens the door to understanding LLM inference and, consequently, to contributing to more sustainable AI and ML system practices.
\begin{figure*}[!htbp]
    \centering
    \includegraphics[width=\linewidth]{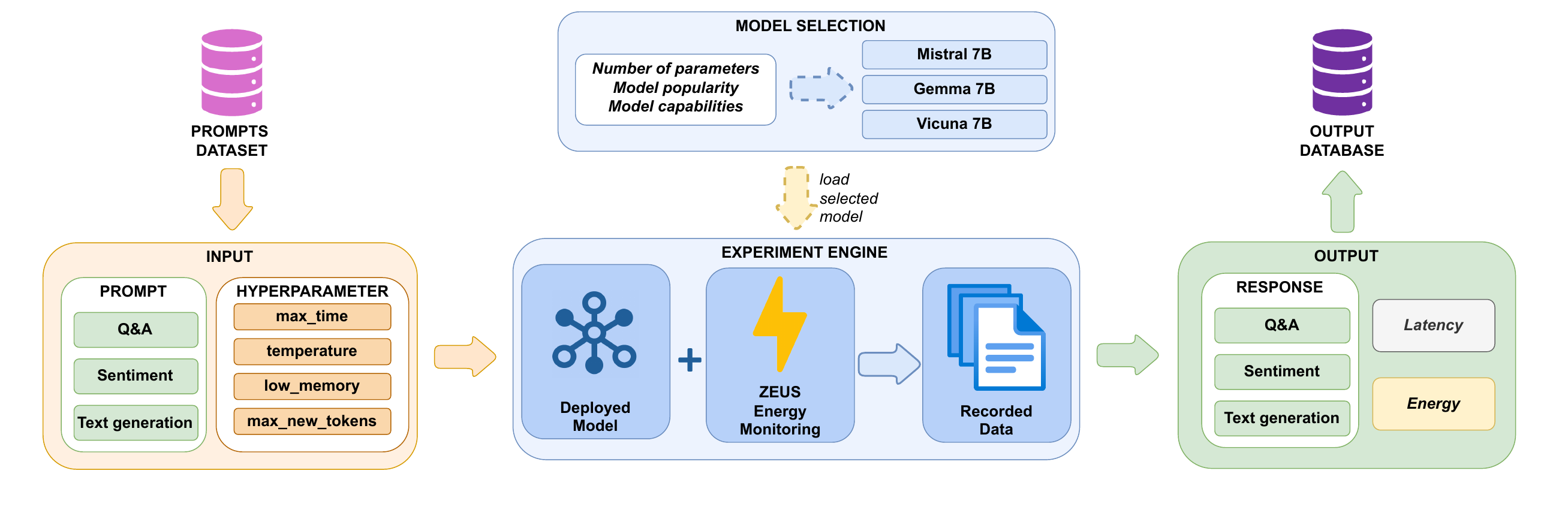}
    \caption{\fontsize{10pt}{12pt}\selectfont Diagram of measurement system designed to conduct the experiments.}
    \label{fig:method}
\end{figure*}
The main contributions of this research are: 
\begin{itemize} 
\item \textbf{Experimental System}: We present a measurement system for capturing energy consumption during one-shot LLM inference operations. This system enables systematic, reproducible energy profiling across diverse production-representative workloads.
\item \textbf{Response Characteristic Analysis}: We analyze generated outputs, quantifying characteristics including token-level response length. Our results demonstrate that response length exhibits a stronger correlation with energy consumption than input prompt length.
\item \textbf{Task and Semantic Impact Analysis}: We demonstrate that task type and prompt semantics significantly influence response generation behavior and consequently the energy consumption. Through systematic experimentation, we establish direct relationships between task complexity and energy utilization. We further investigate how specific prompt tokens affect system response characteristics.
\item \textbf{Energy Reduction Based on Keywords}: We determine that for certain tasks using specific keywords can introduce energy reduction, with some instances reaching 50\% and 60\%.
\end{itemize}

The remainder of this paper is structured as follows. Section \ref{sec:relatedworks} reviews the related work. Details of the experiment are introduced in Section \ref{sec:analysis-design}, while the experiment setup is described in Section \ref{sec:experiment-setup}. Experimental results are given in Section \ref{sec:analysis}. Discussion can be found in Section \ref{sec:discussion}, which is then followed by Section \ref{sec:limits}, describing the limitations of this study. Finally, Section \ref{sec:conclusion} concludes this paper.

\section{Related work} \label{sec:relatedworks}
In recent years, deep neural networks (DNNs) have shown remarkable capabilities across a range of tasks and have attracted broad attention from researchers. Research and development of these networks require significant computational resources, leading to a considerable energy demand. Therefore, designing energy-efficient DNNs deployed in systems is necessary. 

Strubell et al. \cite{strubell2019nlp} quantified the approximate financial and environmental costs of training various LLMs. In the publication, they introduced an approach to compute the total energy consumption of a model by adding up the power consumption of the GPU, CPU, and DRAM and then multiplying this value by the power usage effectiveness (PUE), accounting for the additional energy needed to sustain the entire computing infrastructure \cite{bai2024beyond}. 

Research presented in \cite{jegham_how_2025} describes the impact LLM inference via the lens of energy and water usage, as well as estimated carbon emissions. In \cite{samsi2023words}, the authors studied the computational and energy utilization of inference with LLMs. They used LLaMa 7B, 13B, and 65B as baselines. The experiments were conducted on two different datasets, Alpaca \cite{taori2023stanford} and GSM8K \cite{cobbe2021training}. While the paper benchmarks model's performance, similarly to \cite{jegham_how_2025} it does not explore the reasons behind the observed patterns.


While numerous research efforts have been dedicated to the training phase of LLMs, only a few studies evaluated the inference costs of these models. Canziani et al. \cite{canziani2016analysis} was one of the initial research endeavors that focused on the inference costs. They comprehensively analyzed important metrics such as accuracy, memory footprint, hyperparameters, operations count, inference time, and energy consumption of 14 different DNN models. Li et al. \cite{li2016evaluating} analyzed the patterns in power consumption and energy efficiency of various DNNs and training frameworks. Desislavov et al. \cite{desislavov2023trends} focused their analysis on inference floating point operations (FLOPs) required to process one input item. They studied two representative domains, i.e., Computer Vision (CV) and Natural Language Processing (NLP).

\section{Analysis Design} \label{sec:analysis-design}
The primary objective of this study is to discover characteristics in LLM inference energy usage, specifically focusing on the impact of different types of prompts and hyperparameters. We propose a framework that allows each of the three open-source models to be exposed to an identical set of scenarios, where they are sent prompts with predefined hyperparameter settings and asked to generate a response. For each inference, we record latency and energy used to generate the response, along with the response itself. Figure \ref{fig:method} represents the process and its components: Model Selection step, Experiment engine, Input and Output data.

Upon receiving a prompt, a dedicated program based on ZEUS library was activated to measure the time and energy consumed by the GPU during inference. All data collected during this process was stored in a remote database on a separate machine and accessed via SSH to prevent interference with GPU performance.

\textbf{Model selection:} 
For each experiment, a single model was deployed within the experimental framework, designed to evaluate performance and energy efficiency. To achieve a balance between performance and efficiency, three 7-billion-parameter LLMs were selected for the experiments. This hyperparameter size was specifically chosen to ensure the models could deliver strong performance without exceeding resource constraints. All of the selected models are open-source, built on the Transformer architecture, and fine-tuned for chatbot applications. Despite sharing the same foundational architecture, each model represents a distinct family: Mistral 7B from the Mistral family, Gemma 7B from Gemma, and Vicuna 7B from Llama. 

The models were loaded from HuggingFace. Selection was based on the popularity of the models and their capabilities (i.e., reasoning, question answering, generation). Moreover, it was important that the models were compact enough to run on a single A100 GPU that was available for the entire duration of the experiments. 

\textbf{Input:}
The model’s performance was tested on three categories of input prompts. Based on findings from a preliminary study, the focus was placed on three key tasks that demonstrated both high accuracy and efficient response times: question answering (Q\&A), sentiment analysis, and text generation. For each task, we used three datasets and sampled 1,000 prompts from each dataset to ensure diversity in the tested scenarios.

In order to check more test case scenarios, we created a set of hyperparameters. The hyperparameters provide greater control over the inference/generated response. A prompt combined with a hyperparameter combination is sent to the model following a JSON-format message.

\textbf{Output:}
The collected data consists of model-generated answers paired with their corresponding prompts. To enable thorough analysis, we recorded the energy consumption and inference latency associated with generating each answer. The data is organized by model and the dataset from which each prompt was sourced. Additionally, we saved all relevant hyperparameter settings for reference. We then analyze the recoded data alongside the input and response quality.

\textbf{Experiment engine:}
The experimental setup consists of two primary components: the deployed model and a program that measures the hardware's energy consumption. The energy monitoring program is activated each time a new prompt is sent to the language model and ceases recording upon the completion of the output generation process. All recorded data is subsequently sent to the database for further analysis.

\textbf{Control variables:}
This study investigates what factors influence the energy usage of LLM inference. In order to achieve the most comprehensive results, we study specific control variables that can be highly indicative of influencing energy usage. The control variables were identified via a preliminary study.

\textbf{Models:} To ensure the results reflect broader characteristics of transformer architecture rather than a single model, we conduct experiments on three LLMs. Each model belongs to a different family, introducing subtle architectural variations and unique training datasets. The selected models are as follows: 
\begin{itemize}
    \item \textbf{Mistral 7B} \cite{mistral2023} was developed by MistralAI and performs well on multiple benchmarks. It is fully open source. The model utilizes a sliding attention window and grouped query attention for more efficient inference. Mistral was trained with an 8K context window.

    \item \textbf{Gemma 7B} \cite{gemma2024} is a model from Google, based on their powerful Gemini. It was trained on text, code and mathematical data. The model stands out due to its improved mathematical abilities. Similarly to Mistral, Gemma's context window is 8K tokens.

    \item \textbf{Vicuna 7B} \cite{vicuna2023} is a fine-tuned version of Llama 2 developed by LMSYS. The model was designed with research in mind and is fully open source. It has the smallest context window among the studied models, with a limit of 2048 tokens.
\end{itemize}
\begin{table}[!htbp]
\caption{\fontsize{10pt}{12pt}\selectfont Comparison of model architectures}
\label{tab:model-comparison}
\centering
\begin{tabular}{P{18mm}P{14mm}P{18mm}P{18mm}}
\toprule
\textbf{\centering \multirow{2}{*}{Model}} & 
\textbf{\centering \# Hidden Layers} & 
\textbf{\centering \# Attention Heads} & 
\textbf{\centering Vocabulary Size} \\ \midrule
Gemma 7B            & 28     & 16       & 256000      \\ 
Mistral 7B          & 32     & 32       & 32000       \\ 
Vicuna 7B           & 32     & 32       & 32000       \\ 
\bottomrule
\end{tabular}
\end{table}
\begin{figure*}[!htbp] 
    \centering
    \includegraphics[width=0.9\linewidth]{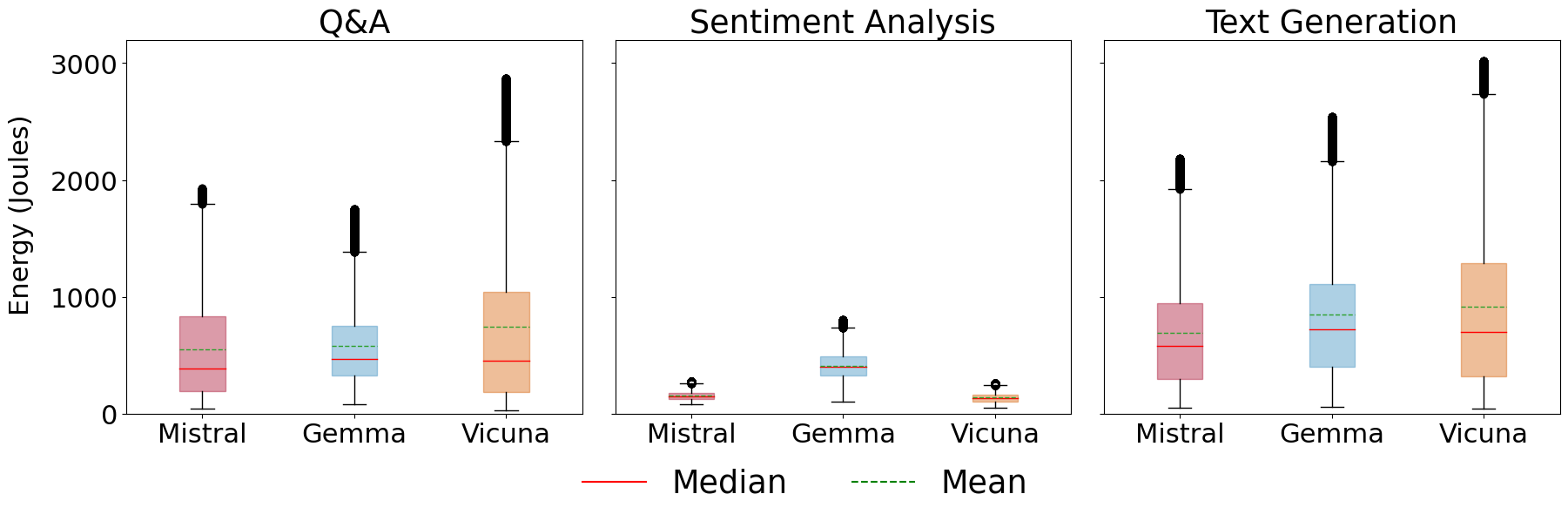}
    \caption{\fontsize{10pt}{12pt}\selectfont Energy consumption measured during inference for each model and task.}
    \label{fig:energy-box}
\end{figure*}
\textbf{Hyperparameters:} Given the constraints of time and computational resources, it was not feasible to explore the entire hyperparameter space. Instead, a subset was selected for evaluation and the values were incrementally adjusted to optimize model performance. For this study, the following hyperparameters were selected for adjustment and each model was evaluated across all hyperparameter settings: 

     \begin{itemize}
         \item \textbf{temperature [0.3, 0.5, 0.7]}: larger temperature settings allow the model to sample from a wider range of possible tokens. This means that responses with higher settings could become more diverse and therefore longer.

         \item \textbf{max\_time [4s, 10s, no limit]}: is used to set a time limit on how long the model is allowed to run during the generation process. It imposes a hard stop on the time the model can spend generating output, regardless of how many tokens have been generated.

         \item \textbf{max\_new\_tokens [10, 50, 100]}: this hyperparameter is used to control the maximum number of tokens that the model is allowed to generate in response to a prompt.

         \item \textbf{low\_memory [True, False]}:  the primary goal is to reduce the amount of memory required by the model, usually by limiting the number of layers or operations that need to be stored in memory at a given time.
\end{itemize}

\textbf{Task:} To gain deeper insights into energy consumption, we evaluated the models across a range of tasks, each characterized by distinct prompt and response length requirements. These variations in task parameters may lead to differing energy usage patterns across tasks. Although the tasks vary, they were selected based on the model's capabilities which include question-answering, sentiment analysis, and text-generation. 
Since most datasets available online focus on a single task type, it was necessary to create a custom dataset by combining prompts from those categories.

\begin{itemize}
    \item \textbf{Q\&A:} category of prompts consists of reading comprehension questions based on a set of Wikipedia articles and context-based instruction-following prompts. This task is composed of SQuAD \cite{squad2018pranav}, GPT4 \cite{gpt2023peng} and WebGLM-QA \cite{webglm2023liu} datasets.

    \item \textbf{Sentiment analysis:} category, using binary classification, includes IMDB movie reviews \cite{imdb2011stanford}, Yelp reviews \cite{yelp2015zhang}, and Twitter posts \cite{twitter2012naji}. It is important to note that for sentiment analysis tasks, we specify that the response should only consist of a JSON-type string containing the sentiment 'positive' or 'negative'. The text generation dataset includes tasks such as text rewriting, summarization, classification, and extraction.

    \item \textbf{Text generation:} task prompts the model to generate text resembling human writing. Prompts of this type often ask to rewrite complex passages or to provide descriptions of events and tend to result in the longest responses. Datasets for this task are OpenOrca \cite{orca2023mukherjee}, HelpSteer\cite{helpsteer2023wang} and GPTeacher General-Instruct\cite{gpteacher}.
\end{itemize}

Using multiple datasets within each task was necessary to enhance data variability. This approach allows for monitoring energy usage under diverse conditions and reduces the influence of bias towards any single data source.
It also helps to ensure that experimental results are not skewed due to features of a single data source.
The final dataset contained 9,000 entries per one hyperparameter combination and included 3,000 prompts from each task category. If available, context was integrated with the prompt.

\begin{figure*}[!htbp]
\centering
  \begin{minipage}{0.49\textwidth}
    \centering
    \includegraphics[width=1\textwidth]{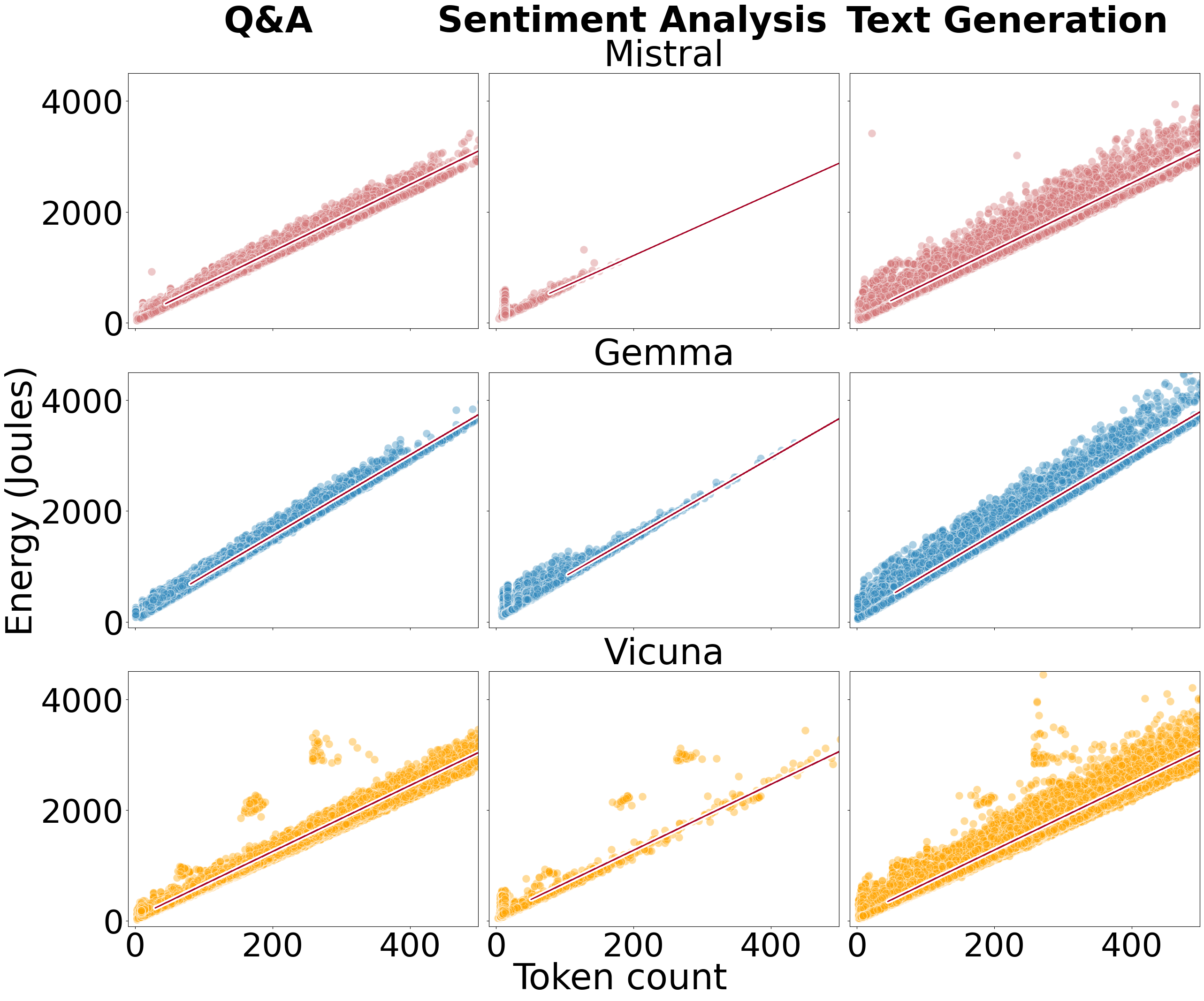}
    \caption{\fontsize{10pt}{10.5pt}\selectfont Scatter plots representing energy usage in relation to response length measured in tokens.}
    \label{fig:scatter-len-energy}
    \end{minipage}
    \hfill
    \begin{minipage}{0.49\textwidth}
    \centering
    \includegraphics[width=1\textwidth]{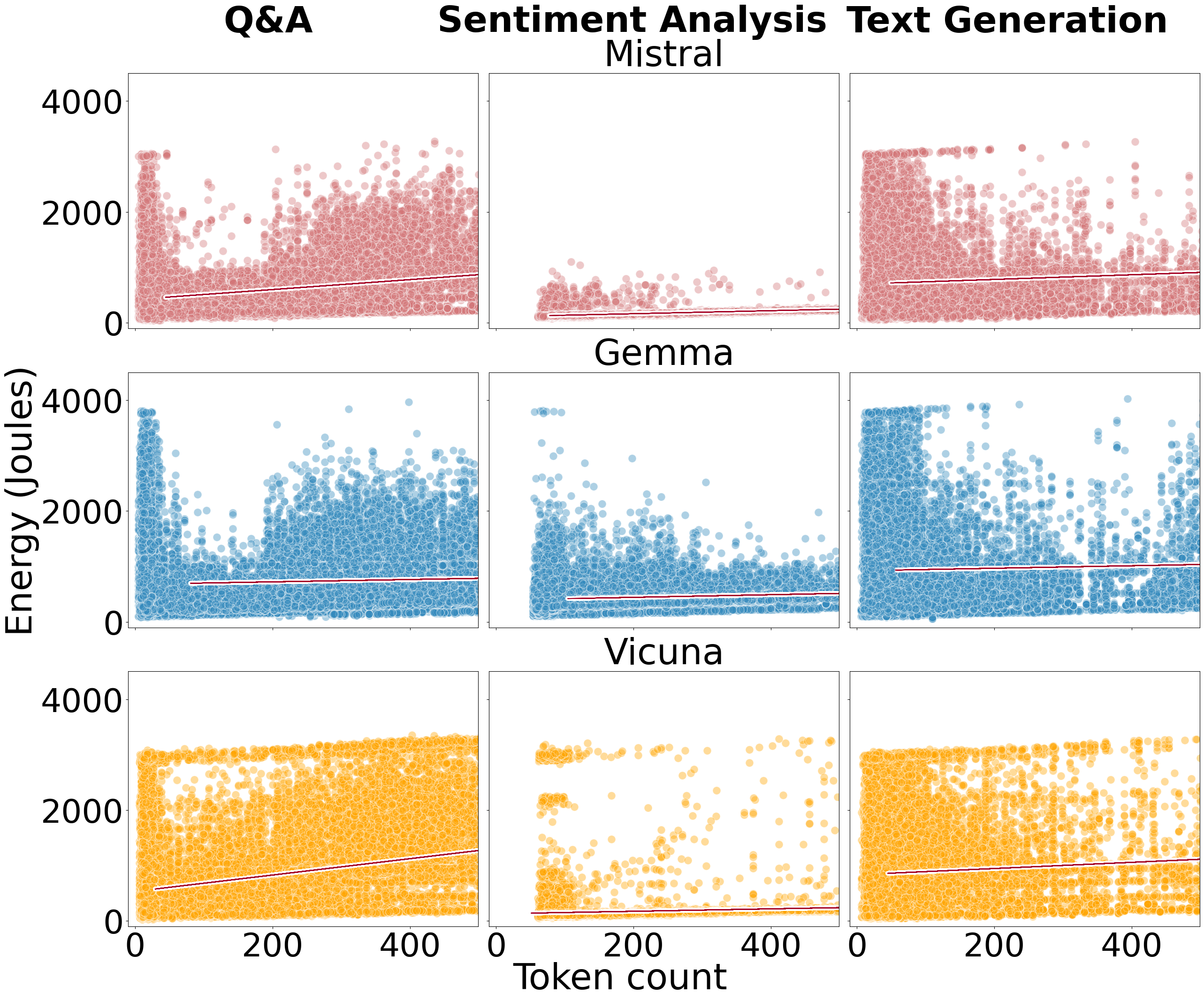}
    \caption{\fontsize{10pt}{10.5pt}\selectfont Scatter plots representing energy usage in relation to prompt length measured in tokens.}
    \label{fig:scatter-prompt-energy}
  \end{minipage}
\end{figure*}

\section{Experiment Setup} \label{sec:experiment-setup}
Prompts were curated and stored in a centralized database. During the experiments, models were deployed one at a time. Each model received prompts, with one prompt being retrieved at a time from the database. The prompt, along with the corresponding hyperparameter settings, was sent to the model in JSON format for processing. This approach ensured that each prompt was handled consistently across all models under identical conditions.

Combining 33 unique LLM model deployment configurations with 9,000 distinct prompts generated a total of 297,000 unique inference requests.

\textbf{Metrics:} By recording both the inference costs and the model configurations, we ensured that detailed comparisons could be made across models, with consistent tracking of performance and resource utilization.

\begin{itemize}
    \item \textbf{Energy consumption:} To conduct a comprehensive study on the impact of prompt characteristics on energy consumption, we systematically collected data on energy usage for each inference. The energy usage was gathered using ZEUS, a service designed to measure GPU energy usage. To ensure the accuracy of our results, we conducted the experiment on a single GPU, carefully controlling the environment by limiting the operation to only the services essential for the experiment. This controlled setup minimized any potential noise variables, allowing for precise measurements of energy consumption across different prompt configurations.

    \item \textbf{Latency:} In addition to measuring energy consumption, we also recorded the time taken by the model to process each prompt, from the moment the input was received to the finalization of the output. 

    \item \textbf{Response length:} Measured in tokens, this metric provides critical insight into the resource demands and performance efficiency of each model under evaluation. The output length as a metric adds a layer of granularity to performance analysis, enabling a more nuanced comparison across model configurations. 
\end{itemize}

\begin{table*}[!ht]
    
    \caption{\fontsize{10pt}{10.5pt}\selectfont Keywords with highest to lowest average energy consumption for tested models, \textit{mean} - represents average energy usage during inference for prompts containing the keyword in Joules, \textit{count} - number of prompts that contain the given keyword, \textit{STD} - standard deviation of energy measured during inference for prompts containing the keyword.}
    \label{tab:model-keywords}

    \centering(a) Mistral 7B
    \vspace{2mm}
    
    \resizebox{\textwidth}{!}{
    \begin{tabular}{P{12mm}P{14mm}P{8mm}P{12mm}P{1mm}P{12mm}P{14mm}P{8mm}P{12mm}P{1mm}P{12mm}P{14mm}P{8mm}P{12mm}}
    \hline
    \multicolumn{4}{c}{Q\&A}               &  & \multicolumn{4}{c}{SENTIMENT ANALYSIS}       &  & \multicolumn{4}{c}{TEXT GENERATION}          \\
    \cline{1-4} \cline{6-9} \cline{11-14} 
    \centering Keyword  & \textit{Mean (J)}    & \textit{Count} & \textit{STD (J)}    &  & \centering Keyword  & \textit{Mean (J)}   & \textit{Count} & \textit{STD (J)}    &  & \centering Keyword  & \textit{Mean (J)}    & \textit{Count} & \textit{STD (J)}     \\
    \toprule
    \textit{justify}  & 1040.57 & 55    & 538.34 &  & \textit{identify}  & 276.96 & 110   & 89.96  &  & \textit{analyse}   & 1618.70 & 33    & 1209.82 \\
    \textit{analyse}  & 950.17  & 11    & 384.34 &  & \textit{classify}  & 272.63 & 11    & 11.38  &  & \textit{recommend} & 1086.18 & 759   & 860.39  \\
    \textit{measure}  & 854.66  & 858   & 516.59 &  & \textit{build}     & 259.65 & 396   & 105.71 &  & \textit{measure}   & 1072.40 & 682   & 793.63  \\
    \textit{create}   & 847.59  & 3146  & 635.12 &  & \textit{explain}   & 258.76 & 517   & 112.35 &  & \textit{report}    & 1050.26 & 1925  & 769.57  \\
    \multicolumn{4}{l}{\vdots}                   &  & \multicolumn{4}{l}{\vdots}                   &  & \multicolumn{4}{l}{\vdots}                     \\
    \textit{summarize}  & 589.14 & 297  & 444.87 &  & \textit{recommend} & 203.21 & 1991  & 70.07  &  & \textit{generate}  & 712.36  & 3608  & 632.56  \\
    \textit{translate}  & 507.14 & 253  & 434.45 &  & \textit{summarize} & 178.08 & 11    & 8.69   &  & \textit{justify}   & 619.76  & 2981  & 554.13  \\
    \textit{write}      & 498.88 & 2310 & 575.62 &  & \textit{analyse}   & 169.36 & 33000 & 65.83  &  & \textit{classify}  & 514.57  & 671   & 593.24  \\
    \textit{classify}   & 312.62 & 352  & 360.65 &  & \textit{provide}   & 169.36 & 33000 & 65.83  &  & \textit{translate} & 482.65  & 1452  & 577.78  \\  
    \bottomrule                                                     
    \end{tabular}}

    \vspace{3mm}
    \centering(b) Gemma 7B
    \vspace{2mm}

    \resizebox{\textwidth}{!}{
    \begin{tabular}{P{12mm}P{14mm}P{8mm}P{12mm}P{1mm}P{12mm}P{14mm}P{8mm}P{12mm}P{1mm}P{12mm}P{14mm}P{8mm}P{12mm}}
    
    \hline
    \multicolumn{4}{c}{Q\&A}               &  & \multicolumn{4}{c}{SENTIMENT ANALYSIS}       &  & \multicolumn{4}{c}{TEXT GENERATION}          \\
    \cline{1-4} \cline{6-9} \cline{11-14} 
    \centering Keyword  & \textit{Mean (J)}    & \textit{Count} & \textit{STD (J)}    &  & \centering Keyword  & \textit{Mean (J)}   & \textit{Count} & \textit{STD (J)}    &  & \centering Keyword  & \textit{Mean (J)}    & \textit{Count} & \textit{STD (J)}     \\
    \toprule
    \textit{justify}    & 1190.35 & 55    & 828.73 &  & \textit{classify} & 592.44 & 11     & 169.31 &  & \textit{analyse}   & 1481.02 & 33    & 783.95 \\
    \textit{explain}    & 937.99  & 1793  & 680.39 &  & \textit{justify}  & 546.97 & 33     & 174.56 &  & \textit{write}     & 1236.50 & 6193  & 997.69 \\
    \textit{create}     & 882.19  & 3146  & 703.64 &  & \textit{identify} & 530.60 & 110    & 175.00 &  & \textit{explain}   & 1229.17 & 4191  & 860.36 \\
    \textit{summarize}  & 857.07  & 297   & 652.69 &  & \textit{explain}  & 508.90 & 517    & 200.33 &  & \textit{recommend} & 1152.60 & 759   & 865.12 \\
    \multicolumn{4}{l}{\vdots}                     &  & \multicolumn{4}{l}{\vdots}                   &  & \multicolumn{4}{l}{\vdots}                    \\
    \textit{review}     & 671.66  & 693   & 568.30 &  & \textit{translate} & 447.70 & 55    & 110.76 &  & \textit{justify}   & 875.81  & 2981  & 614.45 \\
    \textit{report}     & 648.00  & 1496  & 518.90 &  & \textit{analyse}   & 443.09 & 33000 & 222.65 &  & \textit{generate}  & 820.98  & 3608  & 701.98 \\
    \textit{translate}  & 594.33  & 253   & 532.41 &  & \textit{provide}   & 443.09 & 33000 & 222.65 &  & \textit{translate} & 790.62  & 1452  & 682.50 \\
    \textit{classify}   & 500.83  & 352   & 427.33 &  & \textit{summarize} & 438.93 & 11    & 241.45 &  & \textit{classify}  & 766.77  & 671   & 559.50 \\
    \bottomrule
    \end{tabular}}

    \vspace{3mm}
    (c) Vicuna 7B
    \vspace{2mm}
    
    \resizebox{\textwidth}{!}{
    \begin{tabular}{P{12mm}P{14mm}P{8mm}P{12mm}P{1mm}P{12mm}P{14mm}P{8mm}P{12mm}P{1mm}P{12mm}P{14mm}P{8mm}P{12mm}}
    \hline
    \multicolumn{4}{c}{Q\&A}               &  & \multicolumn{4}{c}{SENTIMENT ANALYSIS}       &  & \multicolumn{4}{c}{TEXT GENERATION}          \\
    \cline{1-4} \cline{6-9} \cline{11-14} 
    \centering Keyword  & \textit{Mean (J)}    & \textit{Count} & \textit{STD (J)}    &  & \centering Keyword  & \textit{Mean (J)}   & \textit{Count} & \textit{STD (J)}    &  & \centering Keyword  & \textit{Mean (J)}    & \textit{Count} & \textit{STD (J)}     \\
    \toprule
    \textit{analyse}   & 1745.43 & 11    & 1176.18 &  & \textit{identify}  & 313.28 & 110   & 420.13 &  & \textit{analyse}   & 1641.44 & 33    & 1391.10 \\
    \textit{justify}   & 1481.03 & 55    & 932.86  &  & \textit{build}     & 296.53 & 396   & 385.25 &  & \textit{report}    & 1343.15 & 1925  & 1012.98 \\
    \textit{measure}   & 1241.66 & 858   & 1004.94 &  & \textit{translate} & 278.89 & 55    & 451.89 &  & \textit{measure}   & 1303.94 & 682   & 997.37  \\
    \textit{recommend} & 1194.01 & 506   & 971.83  &  & \textit{explain}   & 263.94 & 517   & 276.24 &  & \textit{recommend} & 1266.19 & 759   & 1089.04 \\
    \multicolumn{4}{l}{\vdots}                    &  & \multicolumn{4}{l}{\vdots}                   &  & \multicolumn{4}{l}{\vdots}                     \\
    \textit{translate} & 832.52  & 253   & 985.83 &  & \textit{measure}   & 186.90 & 66    & 32.97  &  & \textit{justify}   & 920.79  & 2981  & 751.47  \\
    \textit{summarize} & 808.72  & 297   & 805.23 &  & \textit{analyse}   & 181.86 & 33000 & 282.57 &  & \textit{identify}  & 913.32  & 1441  & 738.02  \\
    \textit{write}     & 718.13  & 2310  & 832.41 &  & \textit{provide}   & 181.86 & 33000 & 282.57 &  & \textit{classify}  & 718.12  & 671   & 750.47  \\
    \textit{classify}  & 560.62  & 352   & 767.94 &  & \textit{summarize} & 158.67 & 11    & 6.45   &  & \textit{translate} & 706.52  & 1452  & 815.36  \\
    \bottomrule                                                    
    \end{tabular}}

\end{table*}
\vspace{2mm}

\textbf{Analysis setup:} For the main experiments, a dataset of prompts was used to systematically evaluate the models. Each prompt was processed sequentially, with the time taken for execution and the GPU energy consumption recorded for each instance. The cost of inference was measured using the ZEUS library, which utilizes NVML to get reliable results for NVIDIA GPU energy consumption. The responses generated by the models were stored alongside the performance metrics, enabling a comprehensive analysis of both computational efficiency and model output quality.

\textbf{Energy Measurement:} For energy consumption measurements, the ZEUS library\cite{zeus2023jie} was employed. ZEUS utilizes the NVIDIA Management Library (NVML), providing reliable energy usage data for NVIDIA GPUs. The program recorded both inference time and energy consumption.

\textbf{Data Collection:} 
After processing each prompt, we collect the model's response, energy measured and inference latency. This data, alongside the prompt, hyperparameter settings, and model and prompt dataset name, was logged into the database. This structured data collection allowed for accurate, model-specific energy analysis during post-experiment assessments.

\textbf{Computational resources:} Experiments ran on a local Ubuntu 20.04.6 cluster with a setup made up of Intel Xeon Gold 6336Y (96 cores) at 2 GHz, two NVIDIA A100 GPUs, and 128 GB of RAM. While the system is comprised of two A100s, we only use one to ensure correct energy isolation and eliminate noise during the experiments. 

\section{Analysis Results} \label{sec:analysis}
Our main goal is to evaluate factors influencing the energy consumption of inference for three LLMs. In the evaluation, we focus on the cost—energy and time—as well as word and token count. Additionally, we pay attention to distinctive elements such as the presence of emoticons and keywords. We disabled model quantization and left other optimization techniques for future work. While we recognize the importance of correctness and quality of the provided responses (particularly for less subjective tasks), this is beyond the scope of this study.
Each of the tested models was provided with an identical set of prompts using datasets detailed in Section \ref{sec:analysis-design}, with time and energy consumption measured for every inference. This section introduces an analysis of the prompts, resulting responses, and the associated performance metrics.
The initial analysis examined the correlation between latency (response time) and energy usage. 

\textbf{Response length:} Analyzing the energy cost in relation to the response length revealed a strong correlation between the two. As shown in Figure \ref{fig:scatter-len-energy}, longer responses consistently result in higher energy consumption. This observation was further confirmed by a Pearson correlation analysis, which revealed a correlation coefficient of 0.9, indicating a very strong positive relationship. This indicates that response length is one of the most significant factors influencing energy consumption during LLM inference. This is due to the nature of LLMs: as more tokens are generated, the model must perform additional computations at each step to predict and select the next token. This process involves evaluating a probability distribution in the model vocabulary, often represented in the form of log probabilities (logprobs), which represent the confidence of the model for the possible token choice. Models also utilize mechanisms such as self-attention to maintain coherence and relevance in the output. As a result, inference costs scale with sequence length, leading to increased computational demands over time. This is an important baseline for the following analysis, as we explore the factors that might lead models to generate longer, and therefore more energy-intensive, responses.

\textbf{Model:} Each of the models was presented with the same set of prompts. From the results, we notice that each of the models used on average a different amount of energy. Mistral exhibited the lowest average energy consumption of 547 J. On the other hand, on average, Gemma needed the most energy, 727 J, with Vicuna achieving a mean of 700 J. Detailed comparison of the model's energy usage can be seen in Figure \ref{fig:energy-box}. 

All models were executed on the same hardware and received the same set of tasks, ensuring a consistent testing environment that enabled precise and reliable energy consumption measurements while isolating generated responses as the primary factor contributing to variations in energy usage and model performance.

We found two potential reasons for those differences. The first reason is the varying length of the responses. We notice that Gemma is the most verbose among the models. The second reason we identified is per token cost. We observed that on average Gemma needs the least energy per generated token (0.1 J), while Mistral and Vicuna tend to use more energy, 0.11 J and 0.12 J respectively. 
The further reasons for the observed differences can be attributed to differences in the training data \cite{guo2024stop} – models trained on longer texts, such as articles and books, have a tendency to generate longer responses, and predict the <End of Sequence> token later, whereas models trained on social media posts tend towards concise output. Moreover, longer responses can also result from architectural adjustments. Despite all of the selected models being transformers, all of them have unique features – as we can see in Table \ref{tab:model-comparison}, Gemma’s vocabulary size is 8 times larger than both Mistral’s and Vicuna's vocabulary. This allows the model to have a bigger choice of available tokens, therefore resulting in longer responses. Differences in training data have a major impact on the models, but due to confidentiality and lack of reporting, it is not possible to gather detailed data on this aspect of models. 

\textbf{Prompt length:} In contrast to response length, the number of tokens in a prompt does not seem to have a big influence over the energy used during inference. Figure \ref{fig:scatter-prompt-energy} shows the correlation between the prompt length and the energy usage. While there is a weak, positive correlation of 0.08-0.19 between those factors, it is not as strong as the correlation between the performed task and energy usage. 

\begin{figure*}[h] 
    \centering
    \includegraphics[width=\textwidth]{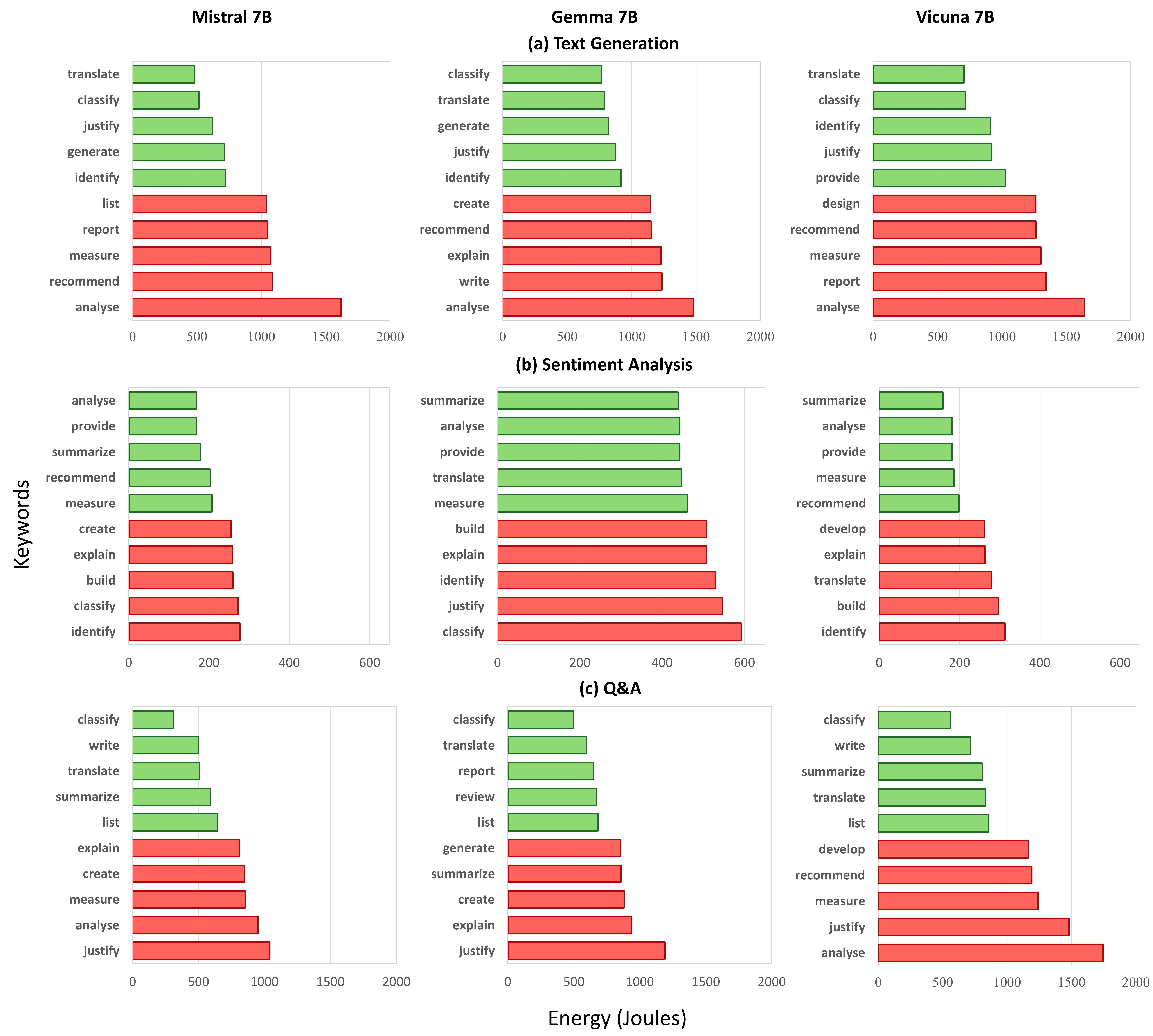}
    \caption{\fontsize{10pt}{12pt}\selectfont Mean Energy Usage (J) per Keyword for Models performing different tasks. Presented keywords are five with the lowest and five with the highest average energy usage.}
    \label{fig:keywords-avg}
    
\end{figure*}

\textbf{Energy per task:} \label{paragraph:task} Distinct variations in energy usage can also be observed across different tasks. As shown in Figure \ref{fig:energy-box}, sentiment analysis required the least energy among all the tasks across the models, with all models remaining below 500 J. Additionally, all models exhibited the lowest variability in energy consumption for sentiment analysis, indicating the most consistent performance across runs.
In contrast, both the Q\&A and text generation tasks demonstrated significant variability in energy consumption, with Vicuna showing the largest deviations between minimum and maximum energy usage in both cases, which can also be noted in Figure \ref{fig:energy-box}. Despite Vicuna’s tendency for larger energy consumption, all three of the models average around 500 J in Q\&A and text generation.
The observed differences in energy usage across tasks can be attributed to the nature of the tasks themselves. For instance, sentiment analysis tends to consume less energy because it typically requires concise responses. In contrast, tasks like Q\&A and text generation often involve more complex demands, such as writing a story or describing a person’s life, which require more verbose and detailed outputs. This shows that the complexity and semantic meaning of the prompt play a significant role in determining the energy consumption of the model when generating a response.

\textbf{Keywords:} \label{paragraph:keywords} To further investigate task-specific differences, we conducted an analysis of keywords, hypothesizing that certain words might correlate with higher energy consumption. We compiled a list of 50 common keywords (e.g., command terms) frequently appearing in our prompt dataset. For each keyword, we calculated the mean energy consumption and its frequency across the database. Table \ref{tab:model-keywords} presents keywords sorted in descending order by the average energy consumption, with three of the highest and lowest mean energy consumption for each model and task.
As shown in the tables, a pattern can be observed in both Q\&A and text generation tasks, which are among the most energy-demanding categories. For instance, the keyword \textit{explain} consistently appears among the top energy-consuming keywords across all models in these tasks – with average 809 J for Mistral, 937 J for Gemma and 1124 J for Vicuna in Q\&A, and 1025 J for Mistral, 1171 J for Gemma and 1124 J for Vicuna in Text Generation.
At the same time, \textit{classify} is positioned at the lower end of the list. In Q\&A it averages 312 J for Mistral, 500 J for Gemma and 560 J for Vicuna. For Text generation, the keyword is associated with prompts that on average use 514 J for Mistral, 767 J for Gemma and 718 J for Vicuna.

For sentiment analysis the keyword \textit{explain} is also among the top contributors to energy consumption, with average 259 J for Mistral, 509 J for Gemma and 264 J for Vicuna.
Keyword \textit{summarize} appears near the bottom, indicating lower energy usage for prompts containing these terms with an average of 178 J for Mistral, 439 J for Gemma and 159 J for Vicuna.

According to our analysis, models share many of the most and least energy-consuming words. The top keywords can be found in Table \ref{tab:model-keywords}. Figure \ref{fig:keywords-avg} shows the difference between the top five most and least energy-consuming keywords for each task. For example, in Text Generation, we can see that for Gemma the difference in average energy usage in prompts can reach up to 50\% between keywords \textit{classify} and \textit{analyse}, while for Mistral this difference is even higher, at 68\%.

Words that correlate with smaller energy usage typically appear in tasks that require more concise responses: \textit{analyse} is commonly seen in sentiment analysis tasks. However, \textit{explain} is much more common in text generation, where the model is asked to provide a detailed description.
These results provide preliminary evidence that certain keywords are associated with higher energy consumption, as they lead to more verbose responses.

\textbf{Emoticons:} We noticed that Vicuna frequently incorporated emoticons in its responses, which prompted us to investigate whether this characteristic impacts energy consumption. We identified all prompts that produced responses with emoticons, revealing that Vicuna included emoticons in over 40\% of its replies, whereas other models used emoticons in less than 1\% of their responses. To assess the potential influence of emoticons on energy consumption, we measured energy usage associated with these responses. The results indicate that, in general, responses containing emoticons do not significantly increase energy demands for Gemma or Mistral, with Pearson correlation nearing 0. However, for Vicuna, the correlation is stronger, with an overall value of about 0.15, but for shorter responses it rises to 0.25. This is because a single emoji is represented by multiple tokens, which drives generation cost. 
Additionally, we noticed a significant number of energy-intensive outliers in Vicuna-generated responses. Upon closer investigation, we found that the model would get stuck and generate a single emoticon until it reached the token limit.

\section{Discussion} \label{sec:discussion}
In this paper, we explored which characteristics of prompts and model responses have the greatest influence on energy consumption during inference. Our findings reveal several important patterns that deserve further examination.

\textbf{Response Length as a Primary Cost Driver:}
We observe that energy usage is tightly correlated with response length, meaning that longer responses use more energy than shorter ones. This finding has practical implications for deployment strategies, as controlling response length can directly reduce inference costs. However, this raises questions about the trade-off between energy efficiency and response quality or its completeness.

\textbf{Semantic Meaning Over Prompt Length:} Notably, the length of a prompt does not directly influence the cost, but it is the semantic meaning that is the key factor. Our results showed that more open-ended tasks, such as text generation, tend to produce longer answers. This tendency comes from the broader scope, lack of specificity and explicitly requested longer content. This observation suggests that prompt engineering for energy efficiency requires attention to the semantic framing of requests. This work opens the door for a deeper discussion on prompt ambiguity and the usage of certain keywords, which might impact the response length as well as the quality.

\textbf{Model-Specific Patterns:}
Despite the models being based on transformer architecture, we can observe different patterns in response length and energy consumption. While Gemma had the lowest cost per token, on average, it produced the longest responses, which led to the highest average cost in total, compared to Mistral and Vicuna. This suggests that training, even for architecturally similar models, can lead to different verbosity tendencies. Models learn from examples in the data where a piece of text naturally ends, helping it predict the end-of-sequence token appropriately. The exploration of this topic is particularly challenging due to the lack of transparency related to training data, which limits our ability to trace the origins of these behavioral differences.

\section{Limitations and Scope} \label{sec:limits}
While this work provides a detailed characterization of prompt-driven energy consumption during LLM inference, it does not propose or evaluate a complete system that dynamically optimizes energy and accuracy. Our focus is intentionally on controlled measurement and analysis, holding hardware, decoding parameters, batching, and model architectures largely constant in order to isolate the impact of prompt semantics on energy usage. We restrict our study to similarly sized dense transformer models. In addition, while accuracy is an important consideration in production deployments, this study does not explicitly couple accuracy metrics with energy consumption, as our objective is to understand the physical cost of inference itself. These choices are deliberate, as the empirical insights presented here are intended to serve as a foundation for future work on building energy-aware systems and architectures and on accuracy–energy analysis.

\section{Conclusion} \label{sec:conclusion}
This study demonstrates that energy consumption during LLM inference is primarily driven by response length rather than prompt length, with the semantic meaning of prompts playing a crucial role in determining overall cost. Our analysis of Mistral 7B, Gemma 7B, and Vicuna 7B reveals that architecturally similar models exhibit distinct energy consumption patterns due to differences in verbosity, highlighting that per-token efficiency does not necessarily translate to lower total energy costs. Open-ended prompts and certain keywords significantly influence response length and subsequent energy usage, suggesting that strategic prompt engineering can serve as an effective approach for reducing inference costs. These findings underscore the importance of considering both model selection and prompt design in the pursuit of energy-efficient LLM deployment, while also revealing the need for further research into the relationship between prompt semantics, response quality, and energy consumption.

This research provides a foundation for investigating the relationship between model complexity and operational cost. As the global carbon footprint of AI scales, understanding the energy cost of every interaction is no longer optional; it is a critical part for the future development of those systems. Our findings demonstrate that the semantic meaning of a prompt is an important driver of inference cost, suggesting that efficiency can be optimized at the user-input level. While this study establishes that keyword selection significantly influences prompt and response length, a more in-depth characterization of the energy profiles associated with specific tokens remains a task for future work. A systematic investigation into these energy-intensive keywords is essential to determine if computational efficiency causes a trade-off in output quality. Finally, these findings provide a framework for developing predictive models capable of estimating energy consumption based on the structural and semantic characteristics of user input.

\section{Acknowledgments}
This work was supported by the Engineering and Physical Sciences Research Council (Fellowship number EP/V007092/1).

\bibliography{bib}

@article{vaswani2017attention,
  title="Attention Is All You Need", 
  author="Ashish Vaswani and Noam Shazeer and Niki Parmar and Jakob Uszkoreit and Llion Jones and Aidan N. Gomez and Lukasz Kaiser and Illia Polosukhin",
  journal="Advances in Neural Information Processing Systems",
  year="2017"
}

@inproceedings{hu2024characterization,
  title="Characterization of large language model development in the datacenter",
  author="Hu, Qinghao and Ye, Zhisheng and Wang, Zerui and Wang, Guoteng and Zhang, Meng and Chen, Qiaoling and Sun, Peng and Lin, Dahua and Wang, Xiaolin and Luo, Yingwei and others",
  booktitle="21st USENIX Symposium on Networked Systems Design and Implementation (NSDI 24)",
  pages="709--729",
  year="2024"
}

@article{khan2021npe,
  title="NPE: An FPGA-based overlay processor for natural language processing",
  author="Khan, Hamza and Khan, Asma and Khan, Zainab and Huang, Lun Bin and Wang, Kun and He, Lei",
  journal="arXiv preprint arXiv:2104.06535",
  year="2021"
}

@misc{jegham_how_2025,
	title = {How {Hungry} is {AI}? {Benchmarking} {Energy}, {Water}, and {Carbon} {Footprint} of {LLM} {Inference}},
	url = {http://arxiv.org/abs/2505.09598},
	doi = {10.48550/arXiv.2505.09598},
	publisher = {arXiv},
	author = {Jegham, Nidhal and Abdelatti, Marwan and Koh, Chan Young and Elmoubarki, Lassad and Hendawi, Abdeltawab},
	year = {2025},
	note = {arXiv:2505.09598 [cs]},
}

@article{armeniakos2022hardware,
  title="Hardware approximate techniques for deep neural network accelerators: A survey",
  author="Armeniakos, Giorgos and Zervakis, Georgios and Soudris, Dimitrios and Henkel, J{\"o}rg",
  journal="ACM Computing Surveys",
  volume="55",
  number="4",
  pages="1--36",
  year="2022",
  publisher="ACM New York, NY"
}

@inproceedings{agrawal2024taming,
  title="Taming $\{$Throughput-Latency$\}$ tradeoff in $\{$LLM$\}$ inference with $\{$Sarathi-Serve$\}$",
  author="Agrawal, Amey and Kedia, Nitin and Panwar, Ashish and Mohan, Jayashree and Kwatra, Nipun and Gulavani, Bhargav and Tumanov, Alexey and Ramjee, Ramachandran",
  booktitle="18th USENIX Symposium on Operating Systems Design and Implementation (OSDI 24)",
  pages="117--134",
  year="2024"
}

@inproceedings{patel2024splitwise,
  title="Splitwise: Efficient generative llm inference using phase splitting",
  author="Patel, Pratyush and Choukse, Esha and Zhang, Chaojie and Shah, Aashaka and Goiri, {\'I}{\~n}igo and Maleki, Saeed and Bianchini, Ricardo",
  booktitle="2024 ACM/IEEE 51st Annual International Symposium on Computer Architecture (ISCA)",
  pages="118--132",
  year="2024",
  organization="IEEE"
}

@article{wu2022sustainable,
  title="Sustainable ai: Environmental implications, challenges and opportunities",
  author="Wu, Carole-Jean and Raghavendra, Ramya and Gupta, Udit and Acun, Bilge and Ardalani, Newsha and Maeng, Kiwan and Chang, Gloria and Aga, Fiona and Huang, Jinshi and Bai, Charles and others",
  journal="Proceedings of Machine Learning and Systems",
  volume="4",
  pages="795--813",
  year="2022"
}

@misc{iea2024electricity,
  author="IEA",
  year="2024",
  url="https://www.iea.org/reports/electricity-2024",
  urldate="February 2025",
  note="Licence: CC BY 4.0"
}

@article{luccioni2023estimating,
  title="Estimating the carbon footprint of bloom, a 176b parameter language model",
  author="Luccioni, Alexandra Sasha and Viguier, Sylvain and Ligozat, Anne-Laure",
  journal="Journal of Machine Learning Research",
  volume="24",
  number="253",
  pages="1--15",
  year="2023"}

@inproceedings{luccioni2024power,
  title="Power hungry processing: Watts driving the cost of ai deployment?",
  author="Luccioni, Sasha and Jernite, Yacine and Strubell, Emma",
  booktitle="Proceedings of the 2024 ACM conference on fairness, accountability, and transparency",
  pages="85--99",
  year="2024"
}

@misc{microsoft2025bing,
  author="Microsoft",
  url="https://www.bing.com/",
  urldate="February 2025",
  year="2025"
}

@misc{bubeck2023sparks,
  title="Sparks of artificial general intelligence: Early experiments with gpt-4",
  author="Bubeck, Sebastien and Chadrasekaran, Varun and Eldan, Ronen and Gehrke, Johannes and Horvitz, Eric and Kamar, Ece and Lee, Peter and Lee, Yin Tat and Li, Yuanzhi and Lundberg, Scott and others",
  year="2023",
  publisher="ArXiv"
}

@inproceedings{zeng2022extensive,
  title="An extensive study on pre-trained models for program understanding and generation",
  author="Zeng, Zhengran and Tan, Hanzhuo and Zhang, Haotian and Li, Jing and Zhang, Yuqun and Zhang, Lingming",
  booktitle="Proceedings of the 31st ACM SIGSOFT international symposium on software testing and analysis",
  pages="39--51",
  year="2022"
}

@article{vatsal2024survey,
  title="A survey of prompt engineering methods in large language models for different nlp tasks",
  author="Vatsal, Shubham and Dubey, Harsh",
  journal="arXiv preprint arXiv:2407.12994",
  year="2024"
}

@article{cheng2023black,
  title="Black-box prompt optimization: Aligning large language models without model training",
  author="Cheng, Jiale and Liu, Xiao and Zheng, Kehan and Ke, Pei and Wang, Hongning and Dong, Yuxiao and Tang, Jie and Huang, Minlie",
  journal="arXiv preprint arXiv:2311.04155",
  year="2023"
}

@article{thirunavukarasu2023large,
  title="Large language models in medicine",
  author="Thirunavukarasu, Arun James and Ting, Darren Shu Jeng and Elangovan, Kabilan and Gutierrez, Laura and Tan, Ting Fang and Ting, Daniel Shu Wei",
  journal="Nature medicine",
  volume="29",
  number="8",
  pages="1930--1940",
  year="2023",
  publisher="Nature Publishing Group US New York"
}

@inproceedings{arora2022ask,
  title="Ask me anything: A simple strategy for prompting language models",
  author="Arora, Simran and Narayan, Avanika and Chen, Mayee F and Orr, Laurel and Guha, Neel and Bhatia, Kush and Chami, Ines and Re, Christopher",
  booktitle="The Eleventh International Conference on Learning Representations",
  year="2022"
}

@article{stojkovic2024towards,
  title="Towards Greener LLMs: Bringing Energy-Efficiency to the Forefront of LLM Inference",
  author="Stojkovic, Jovan and Choukse, Esha and Zhang, Chaojie and Goiri, Inigo and Torrellas, Josep",
  journal="arXiv preprint arXiv:2403.20306",
  year="2024"
}

@inproceedings{strubell2020energy,
  title="Energy and policy considerations for modern deep learning research",
  author="Strubell, Emma and Ganesh, Ananya and McCallum, Andrew",
  booktitle="Proceedings of the AAAI conference on artificial intelligence",
  volume="34",
  number="09",
  pages="13693--13696",
  year="2020"
}

@inproceedings{strubell2019nlp,
author ="Strubell, Emma and Ganesh, Ananya and Mccallum, Andrew",
year ="2019",
month ="01",
pages ="3645-3650",
title ="Energy and Policy Considerations for Deep Learning in NLP",
}

@inproceedings{guo2024stop,
  title="When to stop? towards efficient code generation in llms with excess token prevention",
  author="Guo, Lianghong and Wang, Yanlin and Shi, Ensheng and Zhong, Wanjun and Zhang, Hongyu and Chen, Jiachi and Zhang, Ruikai and Ma, Yuchi and Zheng, Zibin",
  booktitle="Proceedings of the 33rd ACM SIGSOFT International Symposium on Software Testing and Analysis",
  pages="1073--1085",
  year="2024"
}

@article{canziani2016analysis,
  title="An analysis of deep neural network models for practical applications",
  author="Canziani, Alfredo and Paszke, Adam and Culurciello, Eugenio",
  journal="arXiv preprint arXiv:1605.07678",
  year="2016"
}

@inproceedings{li2016evaluating,
  title="Evaluating the energy efficiency of deep convolutional neural networks on CPUs and GPUs",
  author="Li, Da and Chen, Xinbo and Becchi, Michela and Zong, Ziliang",
  year="2016",
  organization="IEEE"
}

@article{bai2024beyond,
  title="Beyond efficiency: A systematic survey of resource-efficient large language models",
  author="Bai, Guangji and Chai, Zheng and Ling, Chen and Wang, Shiyu and others",
  journal="arXiv preprint arXiv:2401.00625",
  year="2024"
}

@article{desislavov2023trends,
  title="Trends in AI inference energy consumption: Beyond the performance-vs-parameter laws of deep learning",
  author="Desislavov, Radosvet and Mart{\'\i}nez-Plumed, Fernando and Hern{\'a}ndez-Orallo, Jos{\'e}",
  journal="Sustainable Computing: Informatics and Systems",
  volume="38",
  pages="100857",
  year="2023",
  publisher="Elsevier"
}

@inproceedings{samsi2023words,
  title="From words to watts: Benchmarking the energy costs of large language model inference",
  author="Samsi, Siddharth and Zhao, Dan and McDonald, Joseph and Li, Baolin and Michaleas, Adam and Jones, Michael and Bergeron, William and Kepner, Jeremy and Tiwari, Devesh and Gadepally, Vijay",
  booktitle="2023 IEEE High Performance Extreme Computing Conference (HPEC)",
  pages="1--9",
  year="2023",
  organization="IEEE"
}

@misc{taori2023stanford,
  title="Stanford alpaca: An instruction-following llama model",
  author="Taori, Rohan and Gulrajani, Ishaan and Zhang, Tianyi and others",
  year="2023"
}

@article{cobbe2021training,
  title="Training verifiers to solve math word problems",
  author="Cobbe, Karl and Kosaraju, Vineet and Bavarian, Mohammad and others",
  journal="arXiv preprint arXiv:2110.14168",
  year="2021"
}

@inproceedings {zeus2023jie,
  author ="Jie You and Jae-Won Chung and Mosharaf Chowdhury",
  title ="Zeus: Understanding and Optimizing {GPU} Energy Consumption of {DNN} Training",
  booktitle ="20th USENIX Symposium on Networked Systems Design and Implementation (NSDI 23)",
  year ="2023",
  pages ="119--139"
}

@misc{mistral2023,
  title="Mistral 7B", 
  author="Albert Q. Jiang and Alexandre Sablayrolles and Arthur Mensch and Chris Bamford and Devendra Singh Chaplot and Diego de las Casas and Florian Bressand and Gianna Lengyel and Guillaume Lample and Lucile Saulnier and Lélio Renard Lavaud and Marie-Anne Lachaux and Pierre Stock and Teven Le Scao and Thibaut Lavril and Thomas Wang and Timothée Lacroix and William El Sayed",
  year="2023"
}

@misc{gemma2024,
  title="Gemma: Open Models Based on Gemini Research and Technology", 
  author="Gemma Team and Thomas Mesnard and Cassidy Hardin and Robert Dadashi et al.",
  year="2024"
}

@misc{vicuna2023,
  title ="Vicuna: An Open-Source Chatbot Impressing GPT-4 with 90\%* ChatGPT Quality",
  author ="Chiang, Wei-Lin and Li, Zhuohan and Lin, Zi and Sheng, Ying and Wu, Zhanghao and Zhang, Hao and Zheng, Lianmin and Zhuang, Siyuan and Zhuang, Yonghao and Gonzalez, Joseph E. and Stoica, Ion and Xing, Eric P.",
  year ="2023"
}

@article{squad2018pranav,
  author="Pranav Rajpurkar and Robin Jia and Percy Liang",
  title="Know What You Don't Know: Unanswerable Questions for SQuAD",
  journal="CoRR",
  year="2018",
  eprint="1806.03822"
}

@inproceedings{imdb2011stanford,
  author="Maas, Andrew L.  and  Daly, Raymond E.  and  Pham, Peter T.  and  Huang, Dan  and  Ng, Andrew Y.  and  Potts, Christopher",
  title="Learning Word Vectors for Sentiment Analysis",
  booktitle="Proceedings of the 49th Annual Meeting of the Association for Computational Linguistics: Human Language Technologies",
  year="2011",
  pages="142--150"
}

@misc{orca2023mukherjee,
  author="Subhabrata Mukherjee and Arindam Mitra and Ganesh Jawahar and Sahaj Agarwal and Hamid Palangi and Ahmed Awadallah",
  title="Orca: Progressive Learning from Complex Explanation Traces of GPT-4",
  eprint="2306.02707",
  year="2023"
}

@misc{gpt2023peng,
  title="Instruction Tuning with GPT-4", 
  author="Baolin Peng and Chunyuan Li and Pengcheng He and Michel Galley and Jianfeng Gao",
  year="2023",
  eprint="2304.03277"
}

@inproceedings{twitter2012naji,
  title="TSATC: Twitter Sentiment Analysis Training Corpus",
  author="Ibrahim Naji",
  booktitle="thinknook",
  year="2012",
  url="https://huggingface.co/datasets/carblacac/twitter-sentiment-analysis",
  urldate="2024-07-05"
}

@misc{helpsteer2023wang,
  title="HelpSteer: Multi-attribute Helpfulness Dataset for SteerLM", 
  author="Zhilin Wang and Yi Dong and Jiaqi Zeng and Virginia Adams and Makesh Narsimhan Sreedhar and Daniel Egert and Olivier Delalleau and Jane Polak Scowcroft and Neel Kant and Aidan Swope and Oleksii Kuchaiev",
  year="2023",
  url="https://arxiv.org/abs/2311.09528"
}

@misc{webglm2023liu,
  title="WebGLM: Towards An Efficient Web-Enhanced Question Answering System with Human Preferences",
  author="Xiao Liu and Hanyu Lai and Hao Yu and Yifan Xu and Aohan Zeng and Zhengxiao Du and Peng Zhang and Yuxiao Dong and Jie Tang",
  year="2023",
  url="https://arxiv.org/abs/2306.07906"
}

@misc{yelp2015zhang,  
  title="Character-Level Convolutional Networks for Text Classification",
  author="Zhang, Xiang and Zhao, Junbo and LeCun, Yann",
  url="https://arxiv.org/abs/1509.01626",
  year="2015"
}

@misc{gpteacher,
  title="{GPT}eacher {G}eneral-{I}nstruct dataset",
  url="https://huggingface.co/datasets/teknium/GPTeacher-General-Instruct",
  urldate="2024-07-05",
  author="",
  year="2023"
}
\bibliographystyle{IEEEtran}

\end{document}